\documentclass[a4paper, 10pt, conference]{IEEEtran}
\IEEEoverridecommandlockouts
\usepackage{cite}
\usepackage[numbers]{natbib}
\usepackage{amsmath,amssymb,amsfonts}
\usepackage{algorithmic}
\usepackage{graphicx}
\usepackage{textcomp}
\usepackage{multirow}
\usepackage{subcaption}
\usepackage{pgfplotstable}
\usepackage{placeins}
\renewcommand\thetable{\Roman{table}}
\makeatletter
\newcommand*{\rom}[1]{\expandafter\@slowromancap\romannumeral #1@}
\makeatother
\usepackage{amsmath}
\DeclareFontFamily{OT1}{pzc}{}
\DeclareFontShape{OT1}{pzc}{m}{it}{<-> s * [1.190] pzcmi7t}{}
\DeclareMathAlphabet{\mathpzc}{OT1}{pzc}{m}{it}
\DeclareMathAlphabet{\mathcal}{OMS}{cmsy}{m}{n}
\usepackage{xcolor}
\def\BibTeX{{\rm B\kern-.05em{\sc i\kern-.025em b}\kern-.08em
    T\kern-.1667em\lower.7ex\hbox{E}\kern-.125emX}}
\begin{document}

\title{Positive Feature Values Prioritized Hierarchical Redundancy Eliminated Tree Augmented Na\"{i}ve Bayes Classifier for Hierarchical Feature Spaces\\
}

\author{\IEEEauthorblockN{Cen Wan}
\IEEEauthorblockA{\textit{Department of Computer Science and Information Systems} \\
\textit{Birkbeck, University of London}\\
London, United Kingdom \\
cen.wan@bbk.ac.uk}
}

\maketitle

\begin{abstract}
The Hierarchical Redundancy Eliminated Tree Augmented Na\"{i}ve Bayes (HRE-TAN) classifier is a semi-na\"{i}ve Bayesian model that learns a type of hierarchical redundancy-free tree-like feature representation to estimate the data distribution. In this work, we propose two new types of positive feature values prioritized hierarchical redundancy eliminated tree augmented na\"{i}ve Bayes classifiers that focus on features bearing positive instance values. The two newly proposed methods are applied to 28 real-world bioinformatics datasets showing better predictive performance than the conventional HRE-TAN classifier.
\end{abstract}

\begin{IEEEkeywords}
Hierarchical feature spaces, Tree Augmented Na\"{i}ve Bayes, Gene Ontology
\end{IEEEkeywords}

\section{Introduction}
This work addresses the classification task of machine learning. We propose two new types of Hierarchical Redundancy Eliminated Tree Augmented Na\"{i}ve Bayes (HRE-TAN) classifier \cite{WanICML}, namely HRE-TAN-Mix and HRE-TAN+ to cope with data consisting of features being hierarchically structured. In this work, we apply HRE-TAN-Mix and HRE-TAN+ on a set of data that exploit the well-known Gene Ontology (GO) terms as features to describe the pro- or anti-longevity effect of genes. The GO terms are organized by a type of ``is-a'' (or generalization-specification) relationship as a directed acyclic graph, where GO terms being close to the root of DAG bear  more generic definitions of genes and GO terms being close to the leaf of DAG bear more specific definitions of genes. This type of hierarchical relationship is widely used for many other domains of data, such as the well-known WordNet \cite{wordnet1995} -- words are naturally organized by the generalization-specification relationship.

The pre-defined hierarchical relationships between features have already been shown as a type of valuable structural information for many machine learning tasks. Several studies have successfully exploited the hierarchical relationships to conduct feature selection~\cite{Wan2014,SDM18, Pablo2020,SHSEL,LASSOFS2} to reduce the dimensionality of data by removing feature redundancies. Most recently, Da Silva et. al (2020) proposed a new hierarchical feature selection method that prioritizes features bearing positive values and outperformed other hierarchical feature selection methods. Moreover, the pre-defined hierarchical relationship can be exploited as a type of constraint for training regression models \cite{GraphLASSO1,GraphLASSO2,LASSOFS3} and learning Bayesian network classifiers \cite{WanACMBCB2015,PGM2020, HieTAN2022}. In this work, we proposed two new types of Bayesian network classifiers by exploiting features bearing positive values and removing feature redundancies according to pre-defined hierarchical relationships.

This paper is organized as follows. Section 2 briefly reviews the background of hierarchical redundancy, the conventional hierarchical redundancy eliminated tree augmented na\"{i}ve Bayes classifier and the lazy learning paradigm. Section 3 proposes two new hierarchical redundancy eliminated tree augmented na\"{i}ve Bayes classifiers, viz. HRE-TAN-Mix and HRE-TAN+. Section 4 presents the experimental methodology and results. Finally, Section 5 presents conclusions and future research directions.

\begin{figure}[b]
\centering
\includegraphics[width=7cm]{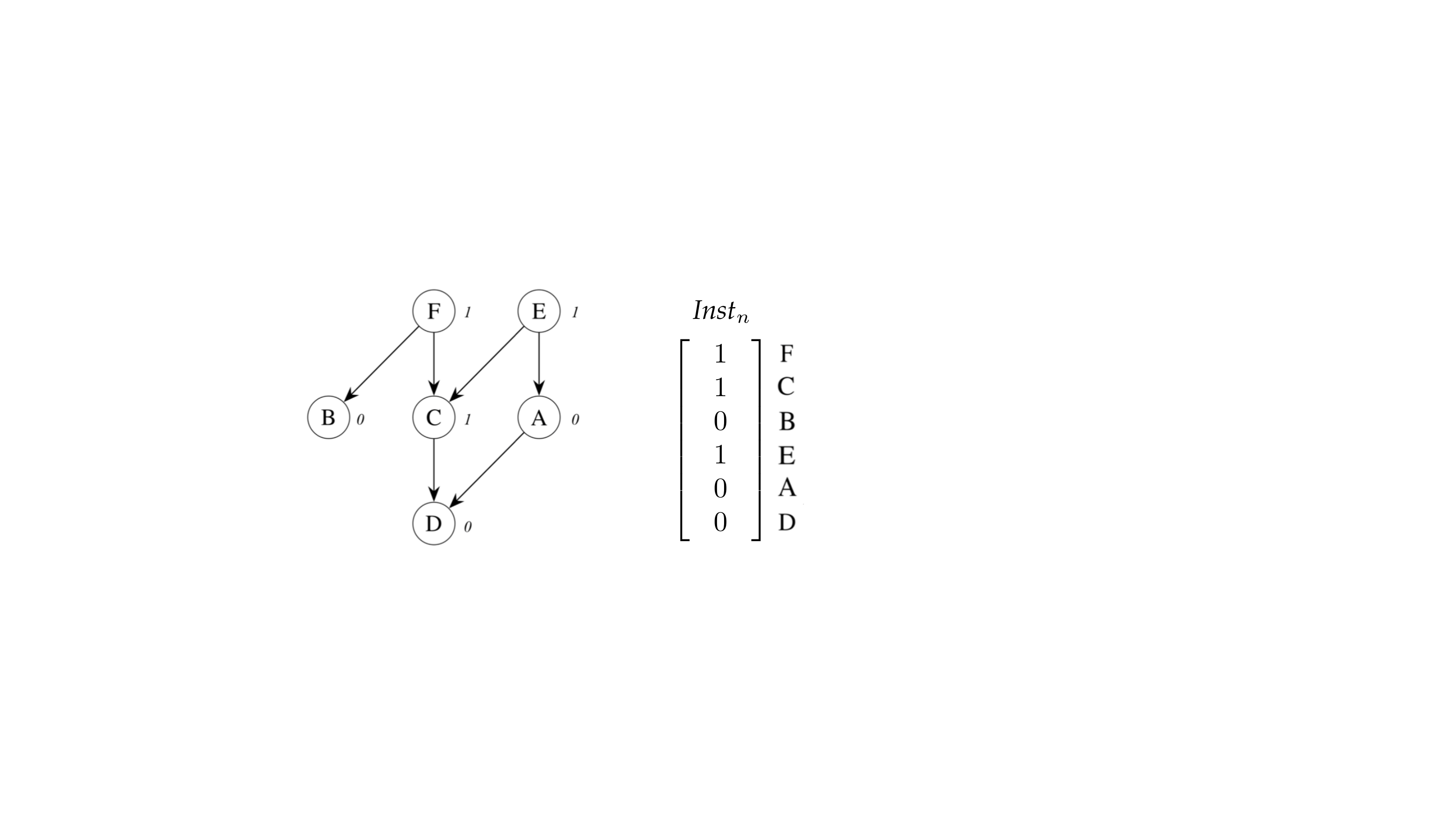}
\caption{An example of feature hierarchy and features' values in an instance.}
\label{fig}
\end{figure}

\section{Background}
\subsection{Hierarchically Structured Features and Hierarchical Redundancy}
Hierarchy is widely used as a type of data structure where different entities are organized by generalization-specialization relationships. In this work, we focus on supervised learning tasks where hierarchically organized entities are used as features. Due to the generalization-specialization relationships, the values of features follow a unique distribution, i.e. if a feature (entity) is annotated with one instance, all its ancestor features (entities) are also annotated with that instance. For example, as shown in Figure 1, six features (entities) F, B, E, C, A, and D are organized as a hierarchy and used to describe an instance (i.e.  $\textit{Inst}_n$). The instance value for feature C equals 1, indicating that the instance values of its ancestors (features F and E) also equal 1. \textit{Vice versa}, if an entity is not annotated with one instance, all its descendant entities are also not annotated with that sample. As feature A's value in $\textit{Inst}_n$ equals 0, the value of its descendant (feature D) in $\textit{Inst}_n$ also equals 0. This type of feature value distribution naturally encodes redundancy between feature values, because the instance values of ancestors or descendant features can be inferred by a single feature's value. For example, the instance values of features F and E can be known by checking the value of their descendant feature C, and the value of feature D can be known by checking the value of its ancestor feature A.

\subsection{Hierarchical Redundancy Eliminated Tree Augmented Na\"{i}ve Bayes Classifier}
Hierarchical Redundancy Eliminated Tree Augmented Na\"{i}ve Bayes (HRE-TAN) \cite{WanICML} is a variant of the well-known Tree Augmented Na\"{i}ve Bayes (TAN) -- a type of semi-na\"{i}ve Bayesian classifier that learns a tree-like feature representation to estimate data distribution. To be different from the conventional TAN, HRE-TAN removes the hierarchical redundancies between features during its training process, i.e. removing edges consisting of at least one hierarchically redundant feature from the set of candidate edges in order to learn a Hierarchical Redundancy Eliminated Maximum Weighted Spanning Tree (HRE-MST). The experimental results confirmed that HRE-TAN significantly outperforms the conventional TAN, and lead to stronger robustness against the class imbalanced issue.

\subsection{Lazy Learning Paradigm}
The lazy learning \cite{Aha1997} paradigm denotes that the classifier training procedure is conducted during the testing phase -- a classifier will be trained and tested for every single testing instance. Analogously to HRE-TAN, our newly proposed HRE-TAN-Mix and HRE-TAN+ methods also follow the lazy learning paradigm, as the features bearing positive values are selected for every single instance.

\section{Proposed Methods}
We proposed two variants of HRE-TAN, i.e. HRE-TAN-Mix and HRE-TAN+. Both newly proposed methods prioritize features that bear positive instance values. The former selects edges including at least one feature bearing positive values as candidate edges whilst the latter only selects edges that consist of pairs of features simultaneously bearing positive values as candidate edges. The pseudocode for HRE-TAN-Mix and HRE-TAN+ is shown in Algorithm 1, whilst the notations to define the proposed HRE-TAN-Mix and HRE-TAN+ are summarize in Table \rom{1}.

\begin{table}[!b]
\centering
\caption{The symbols and notations used in this paper.}
\resizebox{8.9cm}{!}{
\renewcommand{\arraystretch}{1.3}
\begin{tabular}{|c|l|}
\hline
\bf Symbol& \multicolumn{1}{ c| }{\bf Description}\\\hline
\textit{DAG}& the GO hierarchy represented as a Directed Acyclic Graph.\\
\textit{TrainSet}& the training dataset.\\
\textit{TestSet}& the testing dataset.\\
\textit{Inst}$_n$& the $n_{th}$ instance in a testing dataset.\\
$\mathcal{X}$ & the set of all features in a dataset.\\
$\mathcal{A}(x)$ & the set of all ancestors for one feature $x$.\\
$\mathcal{D}(x)$ & the set of all descendents for one feature $x$.\\
$e(x_i, x_j)$ & an edge consisiting of features $x_i$ and $x_j$.\\
$\mathpzc{E}$& the set of edges generated by all features in $\mathcal{X}$.\\
$\mathpzc{E'}$& the set of candidated edges.\\
$\mathcal{S}$& the selection status for all $e(x_i, x_j)$ $\in$ $\mathpzc{E}$.\\
$\mathcal{M}$& the mutual information for all $e(x_i, x_j)$ $\in$ $\mathpzc{E}$.\\
$\mathcal{V}(x)$& the binary feature value for a feature $x$ in $\textit{Inst}_n$.\\
$\mathcal{T}$&the learned tree-like feature presentation by \textit{HRE-MST}.\\
$\mathcal{X'}$ & the set of features included in $\mathcal{T}$.\\
\textit{TrainSet'}& the recreated training dataset by all features in $\mathcal{X'}$.\\
$\textit{Inst'}_n$& the recreated testing instance $n$ by all features in $\mathcal{X'}$.\\
$\mathcal{I}$ & the degree of class imbalance.\\
\hline
\end{tabular}	
}
\end{table}

\begin{figure}[!t]
\centerline{\includegraphics{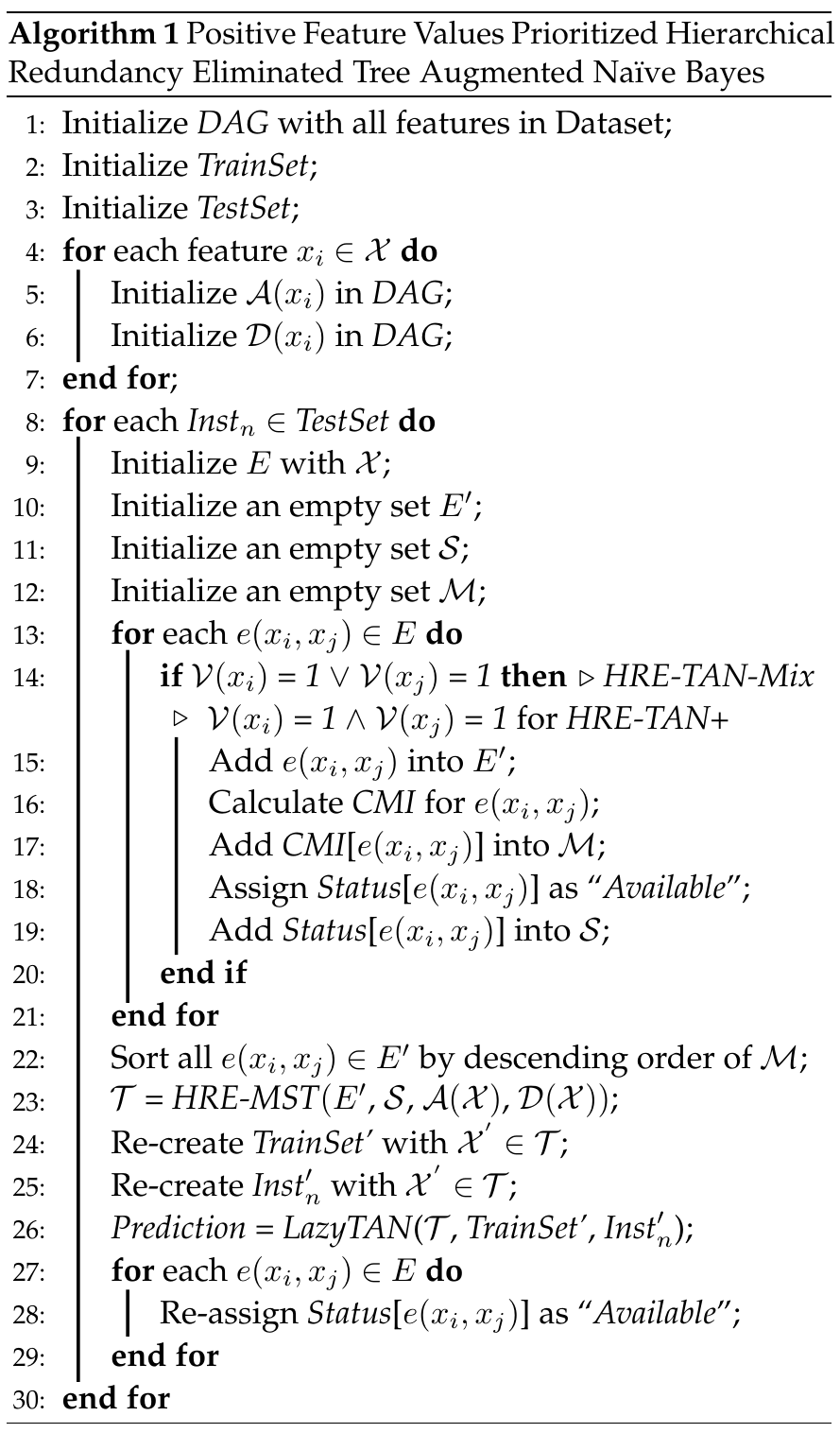}}
\label{fig}
\end{figure}

In Algorithm 1, the first part (lines 1-7) of HRE-TAN-Mix and HRE-TAN+ generates the Directed Acyclic Graph (\textit{DAG}) for all features $\mathcal{X}$, and initializes the training dataset (\textit{TrainSet}) and testing dataset (\textit{TestSet}). Then the ancestor $\mathcal{A}$ and descendant $\mathcal{D}$ sets for individual features $x$ are also initialized by \textit{DAG}. The second part (lines 8-30) of Algorithm 1 conducts the lazy learning procedure of HRE-TAN-Mix and HRE-TAN+ for individual testing instance $\textit{Inst}_{n}$. From lines 9-12, a set of edges $\mathpzc{E}$ that is generated by all possible pairwise combinations of features in $\mathcal{X}$ are initialized. Then three empty sets -- $\mathpzc{E'}$, $\mathcal{S}$ and $\mathcal{M}$ are also initialized, denoting the set of candidate edges, the selection status of candidate edges, and the mutual information of the candidate edges, respectively. Lines 13-21 conduct the preprocessing procedure for learning HRE-TAN-Mix and HRE-TAN+. In terms of HRE-TAN-Mix, for every edge $e(x_i, x_j)$ in $\mathpzc{E}$, if at least one feature (i.e. $x_i$ or $x_j$) has positive value (i.e. $\mathcal{V}(x)$ = 1) in $\textit{Inst}_{n}$, then $e(x_i, x_j)$ will be treated as a candidate edge and added into $\mathpzc{E'}$. Analogously, HRE-TAN+ examines the values of both features $x_i$ and $x_j$ for every edge in $\mathpzc{E}$, and treats those edges including both features simultaneously bearing positive values in $\textit{Inst}_{n}$ as a candidate edges. As shown in lines 16-19, after adding all the candidate edges into $\mathpzc{E'}$, the mutual information of the candidate edges will be calculated and added into $\mathcal{M}$. The selected status of the candidate edges will also be assigned as ``\emph{Available}'', and added into $\mathcal{S}$. Then all candidate edges are sorted in descending order, as shown in line 22. A tree representation of features ($\mathcal{T}$) is learned by the $\textit{HRE-MST}$ procedure \cite{WanICML} by using the candidate edges set ($\mathpzc{E'}$), the selection status ($\mathcal{S}$) of the candidate edges, the ancestor ($\mathcal{A}$) and descendant ($\mathcal{D}$) information for all individual features. From lines 24-26, the learned $\mathcal{T}$ is used to predict the label of $\textit{Inst}_{n}$. Finally, the selection status of all individual edges $e(x_i, x_j)$ are reassigned as ``\emph{Available}'' in order to process the next testing instance.

\begin{figure}[!t]
\centerline{\includegraphics[width=8.5cm]{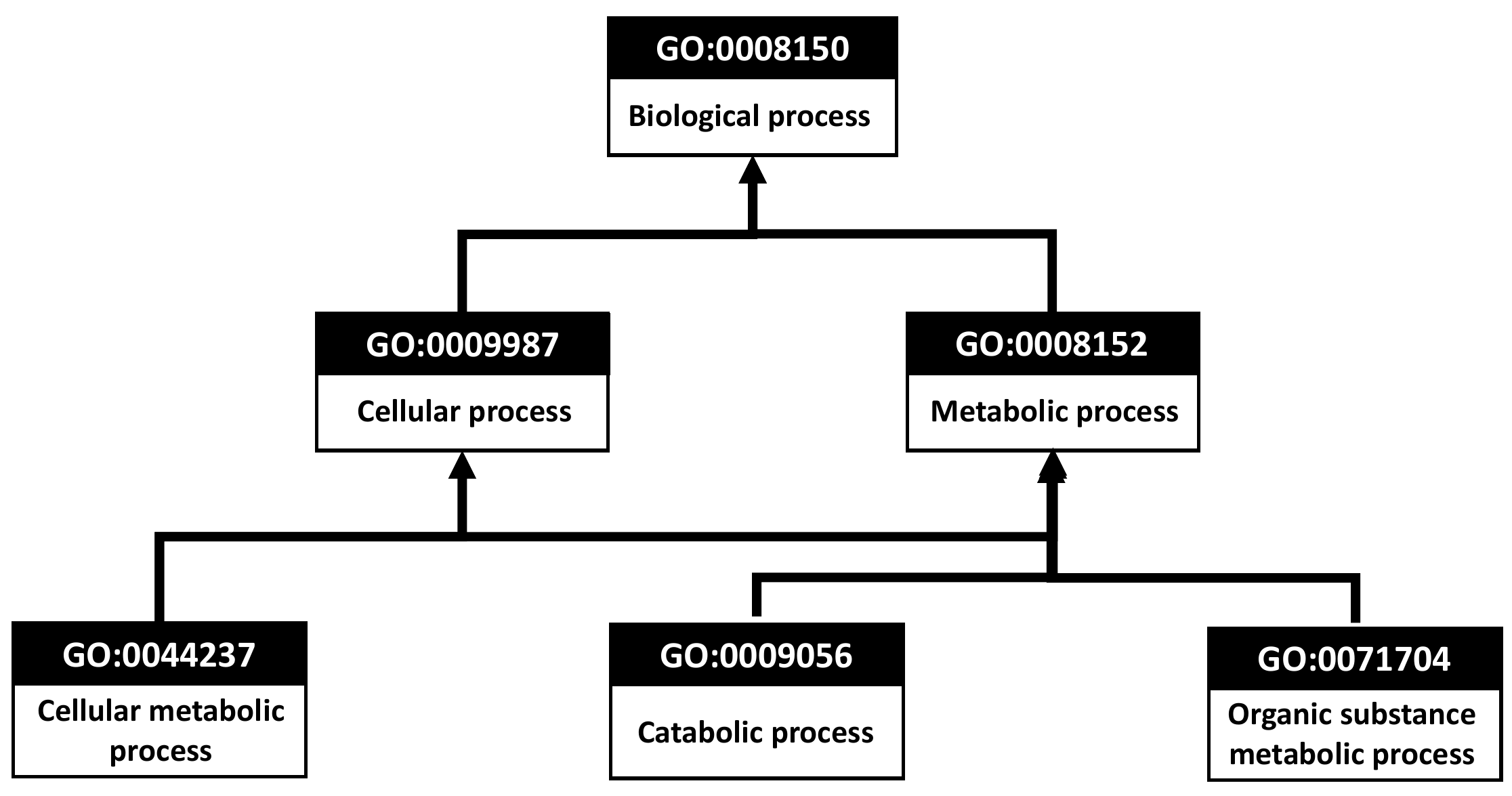}}
\vspace{1.1cm}
\caption{An example of Gene Ontology hierarchy.}
\label{fig}
\end{figure}

\section{Computational Experiments}

\subsection{Experimental Methodology}
We applied the newly proposed HRE-TAN-Mix and HRE-TAN+ methods to predict the effect of genes on aging by using a set of bioinformatics datasets where Gene Ontology \cite{GO2000} terms are used as features to describe model organisms' genes and gene products \cite{CAFA1,CAFA3,FFPredFly}. All Gene Ontology terms are structured by the generalization-specialization relationship, i.e. a GO term that is close to the root of the GO hierarchy denotes a more generic definition of gene function, whereas a GO term that is close to the leaf of the GO hierarchy denotes a more specific definition of gene function. For example, as shown in Figure 2, the term GO:0008150 (biological process) is the root of the example GO hierarchy, denoting the most generic definition of gene function. GO:0009987 (cellular process) and GO:0008152 (metabolic process) are children terms of GO:0008150, denoting two types of more specific biological processes. Those three leaf GO terms denote the most specific definitions of biological process, i.e. term GO:0044237 (cellular metabolic process) denotes a type of more specific cellular process and metabolic process, whilst GO:0009056 (catabolic process) and GO:0071704 (organic substance metabolic process) also denote two types of more specific metabolic process. 

We adopt 28 datasets that were also used in \cite{PGM2020}. Those datasets cover all 572 aging-related genes from four different model organisms, i.e. worm, fly, mouse, and yeast. Each gene is described by seven different types of features, i.e. three individual GO term domains -- biological process (BP), molecular function (MF), and cellular component (CC), and their different types of combinations -- BP+MF, BP+CC, MF+CC, and BP+MF+CC. The dimensions of all 28 datasets range from 75 to 1338. Each gene (instance) is assigned by a class label, i.e. either pro- or anti-longevity.

We use sensitivity, specificity and their geometric mean (GMean) value as metrics to compare the predictive performance of HRE-TAN-Mix and HRE-TAN+ with HRE-TAN. The sensitivity denotes the proportion of correctly predicted samples bearing positive class labels (i.e. pro-longevity), whereas specificity denotes the proportion of correctly predicted samples bearing negative class labels (i.e. anti-longevity). The GMean value is calculated by taking the square root of the product of both sensitivity and specificity, due to the fact that the distributions of the class labels of those 28 datasets are imbalanced. In addition, the well-known 10-fold stratified cross-validation was conducted to obtain the values of those three metrics.

\subsection{Results on Predictive Accuracy}
The computational experimental results are reported in Table \rom{2}, where HRE-TAN-Mix obtained the highest GMean values (in bold) in 15 out of 28 datasets. It also obtained the best overall highest GMean values in the mouse and yeast datasets, i.e. 66.7 and 60.6, respectively. HRE--TAN+ obtained the highest GMean values in 10 out of 28 datasets, whilst obtaining the overall highest Gmean values in the worm and fly datasets, i.e. 61.5 and 64.6, respectively. The benchmark method -- HRE-TAN only obtained the highest GMean values in 4 out of 28 datasets and did not obtain any overall highest GMean value. We further conducted the pairwise Wilcoxon's signed-rank tests on the GMean values obtained by different methods. The significance test results confirm that HRE-TAN-Mix significantly outperforms HRE-TAN+ and HRE-TAN.

\subsection{The robustness against class imbalanced distribution}
We further investigate the robustness of two newly proposed HRE-TAN-Mix and HRE-TAN+ methods against the class imbalanced distribution issue for those 28 datasets. We calculated the Pearson correlation between the GMean values and the degree of class imbalance ($\mathcal{I}$) for each individual dataset. The degree of class imbalance is calculated by taking the difference between 1 and the ratio of samples bearing minority class label over samples bearing majority class label (Equation 1). The values of the class imbalance degree for all 28 datasets are shown in Table \rom{3}. 

\begin{equation}
\mathcal{I}=1-\frac{\mathbf{\textbf{\#}} Minor}{\mathbf{\textbf{\#}} Major}
\end{equation}

\begin{table*}[h]
\caption{Predictive accuracy for HRE-TAN-Mix, HRE-TAN+ and HRE-TAN methods.}
\renewcommand{\arraystretch}{1.45}
\begin{center}
\resizebox{17.6cm}{!}{
\begin{tabular}{|c|ccc|ccc|ccc|}
\hline
\bf Features&\multicolumn{3}{ c| }{\multirow{2}{*}{\bf HRE-TAN-Mix}} & \multicolumn{3}{ c| }{\multirow{2}{*}{\bf HRE-TAN+}}&\multicolumn{3}{ c| }{\multirow{2}{*}{\bf HRE-TAN}}\\
\bf Types      & \multicolumn{3}{ c| }{}  & \multicolumn{3}{ c| }{}                                                              & \multicolumn{3}{ c| }{}        \\\hline
\multicolumn{10}{|c|}{\bf Worm (\emph{Caenorhabditis elegans}) Datasets} \\ \hline
	&	Sens. (SE.)	&	Spec.	(SE.)	&	GMean	&	Sens.	(SE.)	&	Spec.	(SE.)	&	GMean	&	Sens.	(SE.)	&	Spec.	(SE.)	&	GMean	\\\hline
BP	&	34.9	$\pm$	2.7	&	78.3	$\pm$	3.3	&	52.3	&	51.2	$\pm$	3.6	&	71.7	$\pm$	2.5	&	\bf 60.6	&	41.1	$\pm$	2.4	&	76.8	$\pm$	2.1	&	56.2	\\\hline
MF	&	44.6	$\pm$	3.8	&	51.9	$\pm$	5.4	&	\bf48.1	&	21.5	$\pm$	4.5	&	59.5	$\pm$	4.6	&	35.8	&	23.1	$\pm$	4.8	&	75.3	$\pm$	5.4	&	41.7	\\\hline
CC	&	32.0	$\pm$	3.3	&	79.6	$\pm$	2.5	&	\bf50.5	&	16.0	$\pm$	3.7	&	82.0	$\pm$	2.5	&	36.2	&	24.5	$\pm$	3.6	&	80.8	$\pm$	3.0	&	44.5	\\\hline
BP+MF	&	44.6	$\pm$	2.9	&	77.6	$\pm$	2.5	&	58.9	&	55.4	$\pm$	2.0	&	68.2	$\pm$	2.4	&	\bf61.5	&	42.3	$\pm$	2.3	&	80.0	$\pm$	2.6	&	58.2	\\\hline
BP+CC	&	44.6	$\pm$	3.3	&	74.7	$\pm$	2.5	&	\bf57.7	&	43.2	$\pm$	3.1	&	76.1	$\pm$	2.3	&	57.3	&	44.6	$\pm$	3.0	&	74.4	$\pm$	3.6	&	57.6	\\\hline
MF+CC	&	50.6	$\pm$	2.2	&	65.6	$\pm$	3.0	&	\bf57.6	&	28.2	$\pm$	3.9	&	71.4	$\pm$	3.6	&	44.9	&	32.4	$\pm$	3.3	&	79.8	$\pm$	3.2	&	50.8	\\\hline
BP+MF+CC	&	47.9	$\pm$	4.0	&	72.6	$\pm$	2.9	&	59.0	&	49.6	$\pm$	4.4	&	70.7	$\pm$	2.7	&	\bf59.2	&	44.2	$\pm$	3.9	&	79.3	$\pm$	2.9	&	\bf59.2	\\\hline
\multicolumn{10}{|c|}{\bf Fly (\emph{Drosophila melanogaster}) Datasets} \\ \hline			
BP	&	87.8	$\pm$	3.5	&	30.8	$\pm$	10.4	&	52.0	&	83.4	$\pm$	4.8	&	38.3	$\pm$	6.2	&	\bf56.6	&	86.8	$\pm$	3.2	&	30.6	$\pm$	10.2	&	51.5	\\\hline
MF	&	78.0	$\pm$	3.8	&	33.2	$\pm$	8.7	&	50.9	&	54.6	$\pm$	5.2	&	29.8	$\pm$	6.1	&	40.3	&	86.8	$\pm$	3.4	&	41.2	$\pm$	8.8	&	\bf59.8	\\\hline
CC	&	74.2	$\pm$	6.1	&	30.0	$\pm$	7.8	&	\bf47.2	&	75.8	$\pm$	5.6	&	20.0	$\pm$	7.4	&	38.9	&	75.8	$\pm$	5.8	&	28.6	$\pm$	9.7	&	46.6	\\\hline
BP+MF	&	88.9	$\pm$	4.4	&	35.0	$\pm$	8.5	&	55.8	&	82.4	$\pm$	4.8	&	45.0	$\pm$	5.0	&	\bf60.9	&	87.0	$\pm$	3.3	&	31.6	$\pm$	6.5	&	52.4	\\\hline
BP+CC	&	81.3	$\pm$	2.9	&	35.0	$\pm$	9.6	&	53.4	&	83.3	$\pm$	4.5	&	42.5	$\pm$	8.7	&	\bf59.5	&	84.6	$\pm$	2.4	&	32.4	$\pm$	10.6	&	52.4	\\\hline
MF+CC	&	81.3	$\pm$	5.2	&	37.5	$\pm$	8.5	&	55.2	&	76.4	$\pm$	5.0	&	50.0	$\pm$	7.5	&	\bf61.8	&	87.1	$\pm$	4.4	&	39.5	$\pm$	5.5	&	58.7	\\\hline
BP+MF+CC	&	89.1	$\pm$	2.9	&	40.0	$\pm$	8.5	&	59.7	&	83.5	$\pm$	1.9	&	50.0	$\pm$	6.5	&	\bf64.6	&	82.6	$\pm$	3.4	&	47.4	$\pm$	8.7	&	62.6	\\\hline
\multicolumn{10}{|c|}{\bf Mouse (\emph{Mus musculus}) Datasets} \\ \hline	
BP	&	81.7	$\pm$	5.1	&	26.8	$\pm$	6.6	&	46.8	&	80.3	$\pm$	4.2	&	28.5	$\pm$	7.4	&	47.8	&	86.8	$\pm$	5.5	&	47.1	$\pm$	4.7	&	\bf63.9	\\\hline
MF	&	77.1	$\pm$	5.6	&	36.7	$\pm$	8.3	&	53.2	&	75.5	$\pm$	2.4	&	56.7	$\pm$	9.7	&	\bf65.4	&	83.1	$\pm$	3.3	&	42.4	$\pm$	9.3	&	59.4	\\\hline
CC	&	75.2	$\pm$	3.8	&	59.0	$\pm$	10.9	&	\bf66.7	&	70.0	$\pm$	5.4	&	37.6	$\pm$	11.6	&	51.3	&	86.4	$\pm$	4.0	&	41.2	$\pm$	9.7	&	59.7	\\\hline
BP+MF	&	91.4	$\pm$	3.2	&	36.0	$\pm$	7.1	&	57.4	&	85.7	$\pm$	3.0	&	37.3	$\pm$	7.4	&	56.6	&	83.8	$\pm$	4.5	&	41.2	$\pm$	6.8	&	\bf58.8	\\\hline
BP+CC	&	92.9	$\pm$	2.4	&	41.8	$\pm$	8.6	&	\bf62.3	&	83.1	$\pm$	4.5	&	29.8	$\pm$	6.1	&	49.8	&	79.4	$\pm$	4.9	&	47.1	$\pm$	9.7	&	61.2	\\\hline
MF+CC	&	83.7	$\pm$	4.0	&	42.3	$\pm$	8.9	&	59.5	&	79.4	$\pm$	6.1	&	49.3	$\pm$	9.3	&	\bf62.6	&	89.7	$\pm$	3.0	&	35.3	$\pm$	9.6	&	56.3	\\\hline
BP+MF+CC	&	90.0	$\pm$	3.7	&	48.7	$\pm$	10.9	&	\bf66.2	&	82.3	$\pm$	4.1	&	38.7	$\pm$	7.5	&	56.4	&	85.3	$\pm$	3.7	&	44.1	$\pm$	8.9	&	61.3	\\\hline
\multicolumn{10}{|c|}{\bf Yeast (\emph{Saccharomyces cerevisiae}) Datasets} \\ \hline				
BP	&	43.3	$\pm$	11.2	&	83.7	$\pm$	2.4	&	\bf60.2	&	20.0	$\pm$	5.4	&	94.0	$\pm$	1.5	&	43.4	&	20.0	$\pm$	7.4	&	93.5	$\pm$	1.7	&	43.2	\\\hline
MF	&	20.0	$\pm$	8.5	&	83.3	$\pm$	2.9	&	\bf40.8	&	0.0	$\pm$	0.0	&	80.9	$\pm$	2.6	&	0.0	&	0.0	$\pm$	0.0	&	96.9	$\pm$	1.7	&	0.0	\\\hline
CC	&	26.7	$\pm$	9.4	&	83.8	$\pm$	3.0	&	\bf47.3	&	0.0	$\pm$	0.0	&	92.8	$\pm$	1.9	&	0.0	&	12.5	$\pm$	6.1	&	93.5	$\pm$	2.9	&	34.2	\\\hline
BP+MF	&	43.3	$\pm$	10.0	&	84.3	$\pm$	2.2	&	\bf60.4	&	20.0	$\pm$	7.4	&	91.7	$\pm$	1.6	&	42.8	&	26.7	$\pm$	10.9	&	95.8	$\pm$	1.5	&	50.6	\\\hline
BP+CC	&	40.0	$\pm$	6.7	&	90.2	$\pm$	2.0	&	\bf60.1	&	20.0	$\pm$	5.4	&	95.6	$\pm$	1.1	&	43.7	&	26.7	$\pm$	6.7	&	94.1	$\pm$	2.1	&	50.1	\\\hline
MF+CC	&	33.3	$\pm$	11.1	&	84.3	$\pm$	2.9	&	\bf53.0	&	5.0	$\pm$	5.0	&	97.5	$\pm$	0.8	&	22.1	&	10.3	$\pm$	6.1	&	95.4	$\pm$	1.9	&	31.3	\\\hline
BP+MF+CC	&	40.0	$\pm$	8.3	&	91.9	$\pm$	2.0	&	\bf60.6	&	20.0	$\pm$	5.4	&	96.2	$\pm$	1.4	&	43.9	&	23.3	$\pm$	7.1	&	96.2	$\pm$	1.4	&	47.3	\\\hline

\end{tabular}
}
\label{tab1}
\end{center}
\end{table*}
\FloatBarrier

\begin{table}[!t]
\caption{The class imbalance degrees for all 28 datasets.}
\renewcommand{\arraystretch}{1.3}
\resizebox{8.2cm}{!}{
\begin{tabular}{|c|c|c|c|c|}
\hline
\bf Feature & \multirow{2}{*}{\bf Worm} & \multirow{2}{*}{\bf Fly} & \multirow{2}{*}{\bf Mouse} & \multirow{2}{*}{\bf Yeast} \\
\bf Types & & & &\\\hline
BP & 0.345 & 0.604 &0.500 & 0.838 \\\hline
MF & 0.234 & 0.500 & 0.492 & 0.802\\\hline
CC & 0.372 & 0.548 & 0.485 & 0.805\\\hline
BP+MF & 0.374 & 0.587 & 0.500 & 0.844\\\hline
BP+CC & 0.381 & 0.593 & 0.500 & 0.853\\\hline
MF+CC & 0.351 & 0.553 & 0.500 & 0.853\\\hline
BP+MF+CC & 0.398 & 0.587 & 0.500 & 0.856 \\\hline
\end{tabular}	
}
\end{table}

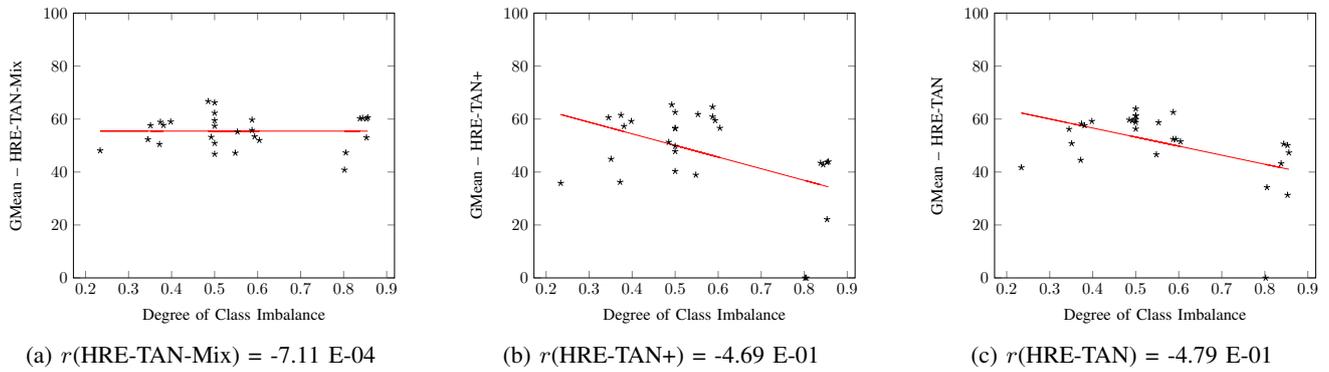
\begin{figure*}[!h]
\centering
  \begin{subfigure}[b]{0.3\textwidth}
    \resizebox{\linewidth}{!}{
     \begin{tikzpicture}
	\begin{axis}[ymin=0, ymax=100,xlabel=Degree of Class Imbalance,ylabel=GMean -- HRE-TAN-Mix]
	\addplot[mark=star,only marks] coordinates {
		(0.345, 52.3)
		(0.234,48.1)
		(0.372,50.5)
		(0.374,58.9)
		(0.381,57.7)
		(0.351,57.6)
		(0.398,59.0)
		(0.604,52.0)
		(0.500, 50.9)
		(0.548,47.2)
		(0.587, 55.8)
		(0.593,53.4)
		(0.553,55.2)
		(0.587, 59.7)
		(0.500, 46.8)
		(0.492, 53.2)
		(0.485,66.7)
		(0.500,57.4)
		(0.500,62.3)
		(0.500,59.5)
		(0.500,66.2)
		(0.838,60.2)
		(0.802,40.8)
		(0.805,47.3)
		(0.844,60.4)
		(0.853,60.1)
		(0.853,53.0)
		(0.856,60.6)
	};
	\addplot table[mark=,row sep=\\, y={create col/linear regression={y=Y}}] 
    {
        X Y\\
0.345 52.3\\
0.234 48.1\\
0.372 50.5\\
0.374 58.9\\
0.381 57.7\\
0.351 57.6\\
0.398 59\\
0.604 52\\
0.5 50.9\\
0.548 47.2\\
0.587 55.8\\
0.593 53.4\\
0.553 55.2\\
0.587 59.7\\
0.5 46.8\\
0.492 53.2\\
0.485 66.7\\
0.5 57.4\\
0.5 62.3\\
0.5 59.5\\
0.5 66.2\\
0.838 60.2\\
0.802 40.8\\
0.805 47.3\\
0.844 60.4\\
0.853 60.1\\
0.853 53\\
0.856 60.6\\
    };
	\end{axis}
   \end{tikzpicture}}
 \subcaption{$r$(HRE-TAN-Mix) = -7.11 E-04}
 \end{subfigure}
\hspace{0.5cm}
  \begin{subfigure}[b]{0.3\textwidth}
    \resizebox{\linewidth}{!}{
     \begin{tikzpicture}
	\begin{axis}[ymin=0, ymax=100,xlabel=Degree of Class Imbalance,ylabel=GMean -- HRE-TAN+]
	\addplot[mark=star,only marks] coordinates {
		(0.345,60.6)
		(0.234,35.8)
		(0.372, 36.2)
		(0.374,61.5)
		(0.381,57.3)
		(0.351,44.9)
		(0.398,59.2)
		(0.604,56.6)
		(0.500,40.3)
		(0.548,38.9)
		(0.587,60.9)
		(0.593,59.5)
		(0.553,61.8)
		(0.587,64.6)
		(0.500, 47.8)
		(0.492,65.4)
		(0.485, 51.3)
		(0.500,56.6)
		(0.500,49.8)
		(0.500,62.6)
		(0.500, 56.4)
		(0.838, 43.4)
		(0.802,0.0)
		(0.805,0.0)
		(0.844, 42.8)
		(0.853,43.7)
		(0.853,22.1)
		(0.856, 43.9)
	};
	\addplot table[mark=,row sep=\\, y={create col/linear regression={y=Y}}] 
    {
        X Y\\
0.345 60.6\\
0.234 35.8\\
0.372 36.2\\
0.374 61.5\\
0.381 57.3\\
0.351 44.9\\
0.398 59.2\\
0.604 56.6\\
0.5 40.3\\
0.548 38.9\\
0.587 60.9\\
0.593 59.5\\
0.553 61.8\\
0.587 64.6\\
0.5 47.8\\
0.492 65.4\\
0.485 51.3\\
0.5 56.6\\
0.5 49.8\\
0.5 62.6\\
0.5 56.4\\
0.838 43.4\\
0.802 0\\
0.805 0\\
0.844 42.8\\
0.853 43.7\\
0.853 22.1\\
0.856 43.9\\
    };
	\end{axis}
   \end{tikzpicture}}
 \subcaption{$r$(HRE-TAN+) = -4.69 E-01}
 \end{subfigure}
\hspace{0.5cm}
 \begin{subfigure}[b]{0.3\textwidth}
    \resizebox{\linewidth}{!}{
     \begin{tikzpicture}
	\begin{axis}[ymin=0, ymax=100,xlabel=Degree of Class Imbalance,ylabel=GMean -- HRE-TAN]
	\addplot[mark=star,only marks] coordinates {
		(0.345,56.2)
		(0.234,41.7)
		(0.372,44.5)
		(0.374,58.2)
		(0.381,57.6)
		(0.351,50.8)
		(0.398,59.2)
		(0.604,51.5)
		(0.500,59.8)
		(0.548,46.6)
		(0.587,52.4)
		(0.593,52.4)
		(0.553,58.7)
		(0.587,62.6)
		(0.500,63.9)
		(0.492,59.4)
		(0.485,59.7)
		(0.500,58.8)
		(0.500,61.2)
		(0.500,56.3)
		(0.500,61.3)
		(0.838,43.2)
		(0.802,0.0)
		(0.805,34.2)
		(0.844,50.6)
		(0.853,50.1)
		(0.853,31.3)
		(0.856,47.3)
	};
	\addplot table[mark=,row sep=\\, y={create col/linear regression={y=Y}}] 
    {
        X Y\\
        0.345 56.2\\
		0.234 41.7\\
		0.372 44.5\\
		0.374 58.2\\
		0.381 57.6\\
		0.351 50.8\\
		0.398 59.2\\
		0.604 51.5\\
		0.500 59.8\\
		0.548 46.6\\
		0.587 52.4\\
		0.593 52.4\\
		0.553 58.7\\
		0.587 62.6\\
		0.500 63.9\\
		0.492 59.4\\
		0.485 59.7\\
		0.500 58.8\\
		0.500 61.2\\
		0.500 56.3\\
		0.500 61.3\\
		0.838 43.2\\
		0.802 0.0\\
		0.805 34.2\\
		0.844 50.6\\
		0.853 50.1\\
		0.853 31.3\\
		0.856 47.3\\
     };   
	\end{axis}
   \end{tikzpicture}}
 \subcaption{$r$(HRE-TAN) = -4.79 E-01}
 \end{subfigure}
\vspace{0.3cm}
 \caption{Linear relationships between degrees of class imbalance and GMean values.}
\end{figure*}

The linear relationships between the class imbalance degrees and GMean values are shown in Figure 3, HRE-TAN-Mix shows the strongest robustness according to its best r-value of -7.11 E-04, suggesting that its performance is almost not affected by the class imbalanced distribution issue. However, both HRE-TAN+ and HRE-TAN show weeker robustness.

\section{Conclusion and Future Research}
We proposed two novel hierarchical redundancy eliminated tree augmented na\"{i}ve Bayes methods that exploit features bearing positive values to construct tree-like feature representations for the classification tasks. The experimental results show that the newly proposed methods outperform the original HRE-TAN method. Future research directions would be to propose other Bayes network classification algorithms that also exploit the features bearing positive values and remove the hierarchical redundancies between features.

\section*{Acknowledgment}
This work is supported by the Birkbeck Research Grant.

\bibliography{mybibfile}
\bibliographystyle{IEEEtranN}

\end{document}